\documentclass[10pt,letterpaper]{article}

\usepackage{cogsci}

\cogscifinalcopy %

\usepackage{graphicx}
\graphicspath{ {./figures/} }
\usepackage{float} %
\usepackage[dvipsnames]{xcolor}
 
\usepackage{hyperref}       %
\usepackage{graphicx}
\usepackage{xcolor}
\newif\ifnotes
\notestrue %
\ifnotes
\newcommand{\mnote}[1]{\textcolor{red}{\textbf{Matthias}: #1}}
\newcommand{\tnote}[1]{\textcolor{blue}{\textbf{Tuan Anh}: #1}}
\else
\newcommand{\mnote}[1]{}
\newcommand{\tnote}[1]{}
\fi

\newcommand\solidrule[1][1cm]{\rule[0.5ex]{#1}{.4pt}}
\newcommand\dashedrule{\mbox{%
  \solidrule[1mm]\hspace{0.5mm}\solidrule[1mm]\hspace{0.5mm}\solidrule[1mm]}}

\usepackage{mathtools}
\newcommand{\given}{\lvert}
\usepackage{amsfonts}       %

\usepackage[acronym,smallcaps,nowarn,section,nogroupskip,nonumberlist]{glossaries}
\makeglossaries
\glsdisablehyper{}
\newacronym{SMC}{smc}{sequential Monte Carlo}
\newacronym{WS}{ws}{wake-sleep}
\newacronym{BWS}{bws}{basic wake-sleep}
\newacronym{RWS}{rws}{reweighted wake--sleep}
\newacronym{MWS}{mws}{memoized wake--sleep}
\newacronym{MCMC}{mcmc}{Markov chain Monte Carlo}
\newacronym{ELBO}{elbo}{evidence lower bound}
\newacronym{VAE}{vae}{variational autoencoder}
\newacronym{IWAE}{iwae}{importance weighted autoencoder}
\newacronym{KL}{kl}{Kullback-Leibler}
\newacronym{NS}{ns}{neuro-symbolic}
\newacronym{BG}{bg}{bigram}
\newacronym{LSTM}{lstm}{long short-term memory}
\newacronym{GP}{gp}{Gaussian process}
\newacronym{S2S}{seq2seq}{sequence-to-sequence}
\newacronym{RBF}{rbf}{radial basis function}
\newacronym{BPL}{bpl}{Bayesian program learning}

\usepackage{pslatex}
\usepackage{apacite}
\usepackage{float} %

\vspace{1cm}
\title{Learning Evolved Combinatorial Symbols with a\\ Neuro-symbolic Generative Model}
 
\author{
    {
    \large 
        \bf Matthias Hofer (mhofer@mit.edu), Tuan Anh Le, Roger Levy, Josh Tenenbaum} \\
      Massachusetts Institute of Technology, Department of Brain and Cognitive Sciences \\
      43 Vassar Street, Cambridge, MA 02143 USA
}

\begin{document}

\maketitle

\begin{abstract}
  Humans have the ability to rapidly understand rich combinatorial concepts from limited data. Here we investigate this ability in the context of auditory signals, which have been evolved in a cultural transmission experiment to study the emergence of combinatorial structure in language.
  We propose a neuro-symbolic generative model which combines the strengths of previous approaches to concept learning.
  Our model performs fast inference drawing on neural network methods, while still retaining the interpretability and generalization from limited data seen in structured generative approaches.
  This model outperforms a purely neural network-based approach on classification as evaluated against both ground truth and human experimental classification preferences, and produces superior reproductions of observed signals as well.
  Our results demonstrate the power of flexible combined neural-symbolic architectures for human-like generalization in raw perceptual domains and offers a step towards developing precise computational models of inductive biases in language evolution.

\end{abstract}

\section{Introduction}
\label{sec:introdution}
The ability to learn and transmit cultural symbol systems is central to human intelligence. From phonemes in words, to notes in a melody or the lines of a diagram, these systems rely on \textit{combinatorial structure}, reusing elements to compose new concepts. 
People learn these systems quickly, and can extend them by learning new concepts from just one or a few examples.
This is not surprising when considering the pressures under which these systems evolved: 
Symbol systems should be both expressive, supporting many distinctions between alternatives, and rapidly learnable, so they can be transmitted easily and robustly \shortcite{smith2013linguistic, kirby2015compression}. 
Combinatoriality appears to be an efficient solution to the trade-off between these factors, %
\shortcite{verhoef2014emergence}, especially when a large repertoire of meanings needs to be expressed \shortcite{nowak1999error}.

The human mind also has a propensity for finding combinatorial patterns in data, even when that structure is not really there.
This is particularly evident in emergent sign language. %
In home sign, deaf children who have no access to spoken language or pre-existing sign languages spontaneously adopt combinatorial patterns in their gestures \shortcite{senghas2004children}. 
Laboratory studies of artificial language learning also show that combinatoriality emerges spontaneously without explicit supervision
\shortcite<e.g.,>{verhoef2014emergence, little2017signal, roberts2015communication, hofer2019iconicity}.

How do such symbol systems develop?
Here we study computational models of the
combinatorial learning and productive reuse %
we see in the these studies to better understand the cognitive processes that enable people to learn and perpetuate these systems.
Existing approaches include both structured probabilistic models \shortcite{feinman2020generating, lake2015human} and neural networks%
~\shortcite{ganin2018synthesizing}. 
In one notable example, \citeA{lake2015human} developed a Bayesian model that acquires handwritten characters through structured motor programs, but it remains unclear how this or similar approaches could be scaled up to other domains besides handwriting, 
or explain how new symbol systems arise.

Here we propose a flexible neuro-symbolic generative model, which combines the distinctive strengths of neural, symbolic, and probabilistic approaches. %
We also work with %
a novel dataset of structured auditory signals, which was collected using methods from iterated learning to simulate the cultural evolution of a symbol system. %
Our model outperforms a straight neural network-based approach on classification and %
signal reproduction tasks based on this dataset, and also behaves in more human-like ways.

\section{Empirical Domain and Dataset}
\label{sec:domain}

\begin{figure*}[!ht]
    \centering
    \includegraphics[width=\textwidth]{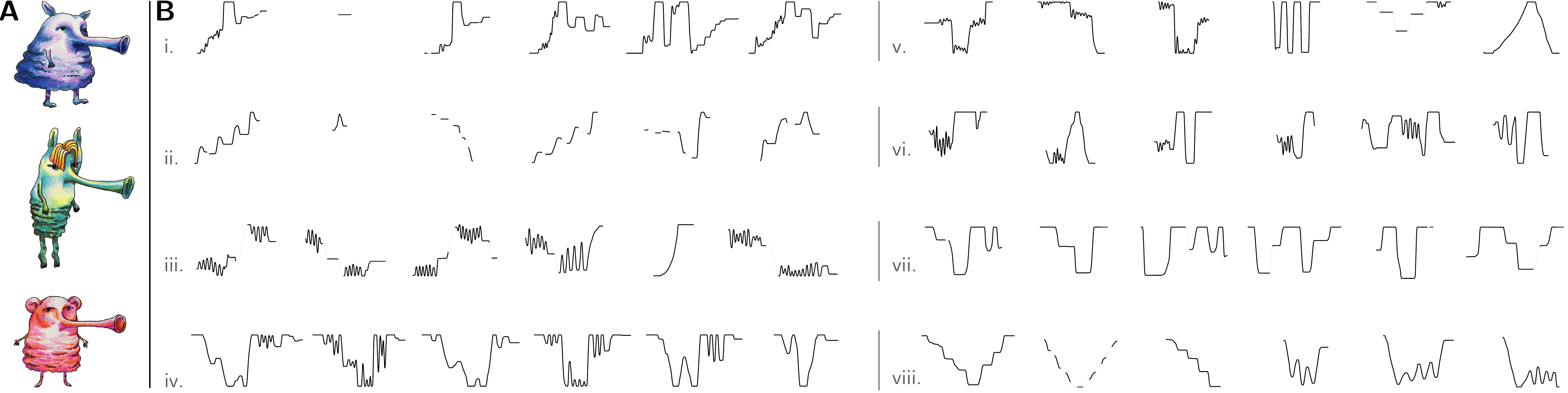}
    \caption{
    Data from \protect\citeA{hofer2019iconicity}.
    (A). Subjects learned whistle-like signals used by aliens on planet \textit{Vuma}. (B). Rows (i) - (viii) show different symbol systems that emerged over the course of the iterated learning experiment, showcasing both stylistic differences and varying degrees of combinatoriality. Signals are visually depicted as pitch trajectories over time (\solidrule[4mm]). The `off` parts of signals where movement but no audible sound occurred are depicted using dashed line segments (\textcolor{gray}{\protect\dashedrule}).
    }
    \vspace{-1em} %
    \label{fig:iterated}
\end{figure*}

Humans already possess rich knowledge of patterns in domains such as writing \shortcite{lake2015human} or speech \shortcite{de2005evolution}---often a lifetime's worth---making it difficult to tease apart contributions of prior bias and learned experience. 
Here we focus on a signal domain where people have little to no prior experience: auditory signals created using a slide whistle instrument \shortcite{verhoef2014emergence, hofer2019iconicity}. Similarly to speech or handwriting, whistled signals are grounded in a perceptually rich substrate, supporting many stylistically different systems, but promise to be more tractable for modeling.

The data we model stem from an iterated learning experiment \cite<see>[for details]{hofer2019iconicity}. Following a relatively standard setup \shortcite<see, e.g.,>{kirby2014iterated}, subjects had to learn and subsequently pass on a small repertoire of whistled signals. Transmission occurred for a total of 10 generations across 15 independent chains. 
Fig.~\ref{fig:iterated}B shows several example languages that evolved in the experiment.
Similarly to home sign alluded to earlier, participants' spontaneous tendency to detect and reuse (ever so spurious) patterns lead to the gradual emergence of combinatorial structure: Signals within a system came to be constructed using shared building blocks (e.g., a rising sweep, a short beep, and a sustained oscillation in system (iii.) in Fig.~\ref{fig:iterated}B).

Why does this structure emerge? One idea is that combinatoriality emerges as an evolutionary trade-off between learnability and communicative utility \shortcite{kirby2015compression}:
While the compression benefits afforded by shared building blocks make these systems more learnable, the ability to systematically rearrange them in novel ways allows users to keep signals distinct. 
Here we take a first step towards testing this hypothesis by specifying a computational model of learning in the slide whistle domain, which, amplified through cultural evolution, could give rise to combinatorial structure.
A key component of the model's inductive bias, which we hypothesize is crucial for systematic generalization from limited data, is an explicit representation of symbols, corresponding to signal parts, that can subsequently be reused in novel contexts. %

\section{Neuro-symbolic generative model}
\label{sec:model}
\begin{figure*}[!ht]
    \centering
    \includegraphics[width=\textwidth]{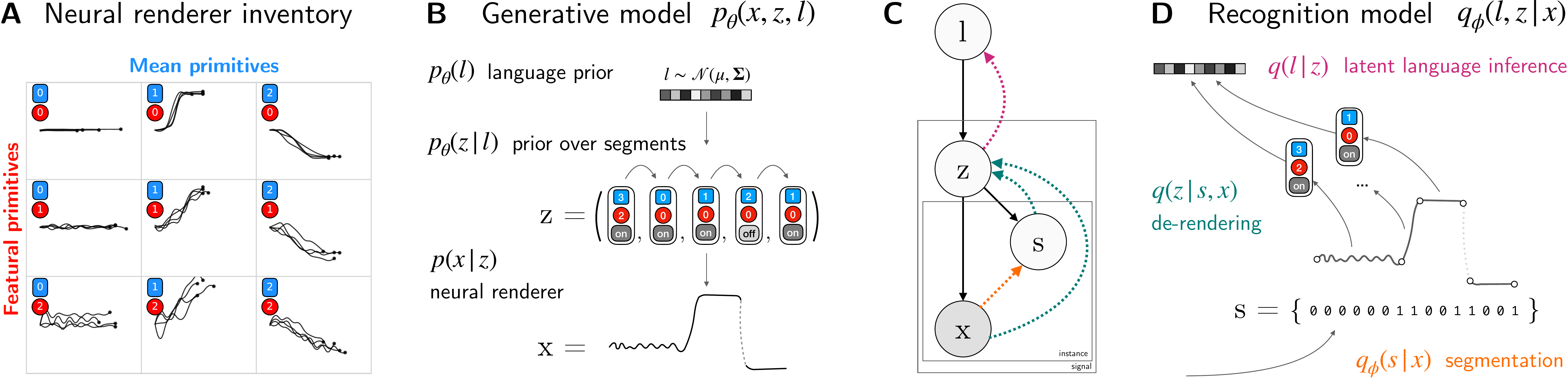}
    \caption{
    \textbf{Neuro-symbolic generative model.}
    Starting from an \emph{inventory} of mean and featural primitives (A), we compose each signal using a \emph{prior} over segments $z \sim p_\theta(z)$ and a \emph{neural renderer} $p(x \given z)$ which produces the continuous signal $x$ (B).
    The \emph{recognition model} is decomposed into \emph{segmentation} $q_\phi(s \given x)$ (either filtering- or neural network-based) and \emph{derendering} $q(z \given s, x)$ (D).
    To model sets of signals, we introduce a global latent code $\ell$ which parameterizes the bigram conditional prior $p_\theta(z \given \ell)$ and a neural recognition model $q_\phi(\ell \given z_{1:M})$ which, together with the single-signal recognition model, fully amortizes inference.
    }
    \vspace{-1em} %
    \label{fig:model}
\end{figure*}

Our \gls{NS} model is a probabilistic generative model with an interpretable symbolic latent variable (see Fig.~\ref{fig:model}C).
Signals are modeled as sequences of discrete segments, which, in turn, are modeled as compositions of primitives.
This sequence of segments is then rendered in the continuous signal-transmission modality via neural networks to produce the observed data.
As such, our model utilizes the kind of systematic generalization afforded by symbolic approaches, as well as the flexibility in bridging to continuous perceptual features afforded by neural components.
Being probabilistic and generative by design, the model naturally supports sampling novel signals from the prior as well as performing Bayesian inference to reproduce and classify signals.
We will first describe our model of single signals $p_\theta(z, x)$ before describing the full hierarchical model of sets of signals, or languages, $p(\ell)\prod_{m = 1}^M p_\theta(x, z \given \ell)$.

\paragraph{Latent sequence prior.}
The latent signal representation $z$ is a sequence of indices of variable length, representing \textit{segments}, which select from a set of primitives (Fig.~\ref{fig:model}B).

Formally, $z$ consists of a sequence of mean--feature--on/off index tuples $i_{1:N} := (i_m^n, i_f^n, i_o^n)_{n = 1}^N$, over which we place a \gls{LSTM}-based auto-regressive prior.
The \gls{LSTM} prior is initialized with zero hidden state, cell state, and input.
At each segment $n$, the \gls{LSTM} cell produces logits of three Categorical distributions over $i_m^n, i_f^n, i_o^n$, which are conditionally independent given the history before step $n$.
The sampled id tuples are then transformed into concatenated one-hot vectors and fed as the input to the \gls{LSTM} at the next step.

\paragraph{Primitives.}
The composition of each segment into a \emph{mean}, a \emph{featural}, and an \emph{on/off} indicator (Fig.~\ref{fig:model}B) exploits the intrinsic organization of the signal-production modality in terms of its distinct articulatory dimensions: the pitch dimension, split into a mean pitch ``contour'' and high frequency (featural) variation around that mean, and the on/off dimension, indicating whether a segment is audible or silent.
Together these primitives form an interpretable, factorized representation of signal elements (e.g., ``rising and wiggly'').\footnote{This hierarchical decomposition signals into segments and segments into primitives mirrors the structure of natural language phonology, which is similarly organized in terms of distinctive features and phonemes.}

\paragraph{Neural renderer-based likelihood.}
An important challenge in model specification is posed by the interface between symbolic and continuous representations. Here we utilize a flexible \gls{LSTM}-based neural network model. 
This \textit{neural renderer} takes as input indices to sets of learnable mean embeddings and feature embeddings.
The two embeddings are then fed as input at every timestep of the \gls{LSTM}, which defines an autoregressive time-series distribution for each mean-feature index tuple as shown in the matrix in Fig.~\ref{fig:model}A.
Signals are obtained by concatenating the time-series segments sampled for each $z_n$ in the latent sequence.

When evaluating $p(x|z)$ there is uncertainty about segmentation $s$ which is required to score the signal segment by segment using the neural renderer. We approximate this quantity by computing $p(x|z) \approx \max_s p(x, s|z)$ using samples of $s$ obtained from the generative model, $p(s | z)$, and from a segmentation model, $q(s | x)$, which is part of the recognition model. %

\subsection{Inference}
Inference is key to learning as well as test-time reproduction and classification.
We use a recognition model (Fig.~\ref{fig:model}D) to propose approximately correct latent sequences and Monte Carlo methods to refine these parses with the generative model in the loop.
Our recognition model is split up into a segmentation model and a derendering model. 

\subsubsection{Segmentation}
Our \textcolor{orange}{\textbf{segmentation model}} is a mixture distribution $q_{\phi}(s|x)=\pi \; q^F(s|x) + (1-\pi) \; q_{\phi}^N(s|x)$, which selects between a filtering-based and a neural segmentation model. %

\paragraph{Filtering-based segmentation.}
One way we generate segmentation proposals is through a multi-scale filtering approach coupled with derivative-based segmentation.
Our approach is similar to existing multi-scale approaches using Gaussian convolutions \shortcite{witkin1984scale}, but instead uses a Gaussian Process-based filter, utilizing a \gls{RBF} kernel to remove higher frequency, featural variation from the signal.
We then use the zero crossings of the differentiated filtered signal as segmentation points $s_{1:N}$.
We place a uniform distribution $q_\phi^F(s \given x)$ over a set of segmentation points, each from a separate setting of the \gls{RBF} prior.

\begin{figure*}[!ht]
    \centering
    \includegraphics[width=\textwidth]{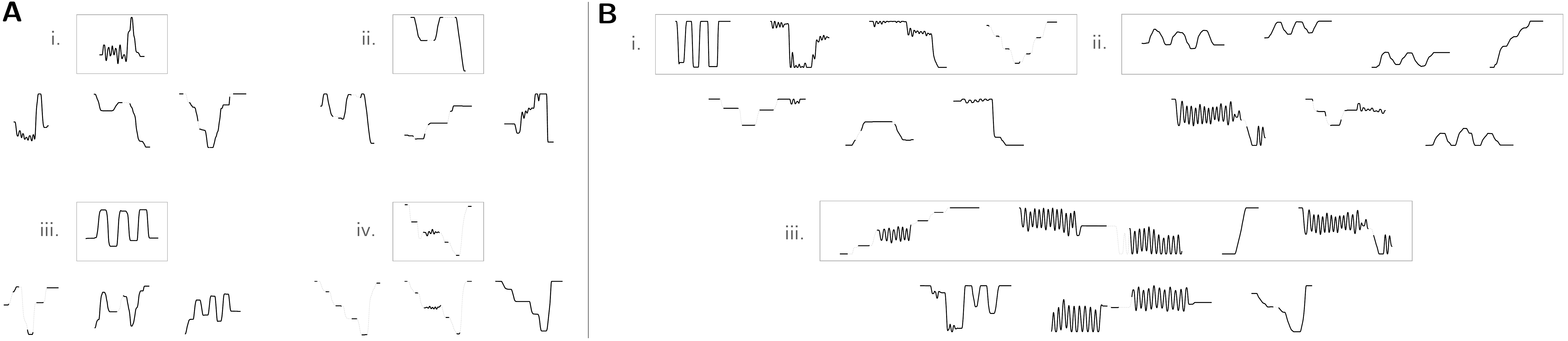}
    \caption{
    (A.) Experimental stimuli used in a three-way signal classification task. The goal is to identify the correct reproduction of the target signal (top) among the three distractors.
    (B.) Stimuli for the language classification task. Here the goal is to pick the signal that belongs to the same set as the target language (top).
    }
    \vspace{-1em} %
    \label{fig:tasks}
\end{figure*}

\paragraph{Neural segmentation.}
We also train a neural segmentation model $q_\phi^N(s \given x)$ as a faster and more general version of filtering-based segmentation. %
We use a \gls{S2S} model that embeds the data $x_{1:T}$ and unrolls an \gls{LSTM} to auto-regressively predict a segmentation vector $s_{1:T}$ where element $s_t$ indicates whether a segment ends at $t$.  %
The input of the decoder \gls{LSTM} at each timestep $t$ is a concatenation of the previous segmentation value $s_{t - 1}$ and the current signal value $x_t$ which provides additional context information.
The \gls{LSTM}'s hidden state is then mapped to a probability of $s_t$.

\subsubsection{Derendering}
Given the segmentation points, we obtain the latent parse $z$ through a neural \textcolor{PineGreen}{\textbf{derendering model} $q(z \given s, x)$}, which was trained to invert the neural renderer and predict latent ids based on signal segments.

\subsubsection{Top-down Monte Carlo inference}
While the recognition model provides fast one-shot inference, we additionally perform importance sampling inference with the top-down generative model in the loop to improve inference.
To do so, we propose a set of samples from the recognition model and weigh them according the ratio of the joint probability and the recognition model probability.

\subsection{Learning}
Learning is split into two parts. 
We first train the neural renderer and derenderer on a dataset of signal means and features (similar to the data depicted in Fig.~\ref{fig:model}A). To obtain this dataset, held-out signals were separated into means and features and subsequently split into segments similarly to the procedure described in the filtering-based segmentation model.
Signal segments were then grouped into categories using unsupervised clustering techniques.

In a second step, we learn the remaining parameters of the generative and recognition models using \gls{MWS} \shortcite{hewitt2020learning}. 
Unlike methods for training deep generative models with continuous latent variables~\shortcite{kingma2014auto}, wake--sleep methods \shortcite{hinton1995wake} learn the generative model and the recognition model in distinct steps which allows more efficient learning in discrete latent-variable models.
Additionally, \gls{MWS} improves learning by maintaining a ``memory'' of best parses so far which stabilizes the parse for each signal from iteration to iteration and reduces the effort spent on search that would otherwise needed to be done by the recognition model from scratch at every iteration.

\subsection{Hierarchical model}
To model languages, or shared structure among sets of signals, we extend the generative model to contain a latent representation $\ell$ which conditions the prior over signal parses $z$ in the language $p_\theta(z \given \ell)$ (Fig.~\ref{fig:model}B).
Given each parse, the signals are generated using the same neural renderer as used in the single signal case.

We choose a standard multivariate Gaussian for the prior $p(\ell)$.
To capture high level language features, we introduce a learnable mapping from $\ell$ to bigram probability matrices which parameterize the transition $P(i_m' \given i_m)P(i_f' \given i_f)P(i_o' \given i_o)$ from the current segment ids $(i_m, i_f, i_o)$ to the next segment ids $(i_m', i_f', i_o')$.
The learnable mapping makes the transition probabilities learnable, and potentially non-Gaussian.

To make inference fully amortized, we train a \textcolor{Magenta}{\textbf{recognition model} $q(\ell \given z_{1:M})$} which conditions on a set of $M$ parses (Fig.~\ref{fig:model}D).
We use a shared \gls{LSTM} to encode each parse and sum up the embeddings before mapping to the mean and variance of a multivariate Gaussian distribution.
For training, we reuse the single signal recognition model to infer a set of parses $z_{1:M}$ for each set of signals $x_{1:M}$.
This reduces the model to a variational autoencoder with inferred $z_{1:M}$ being the data and $\ell$ being the latent `code' which is then trained using evidence lower bound maximization~\shortcite{kingma2014auto}.

\section{Tasks and Baseline Model}
\label{sec:tasks}
To evaluate models, we test their ability to generalize to unseen input from limited training data and compare the results to humans performing the same tasks.

\begin{figure*}[!ht]
    \centering
    \includegraphics[width=\textwidth]{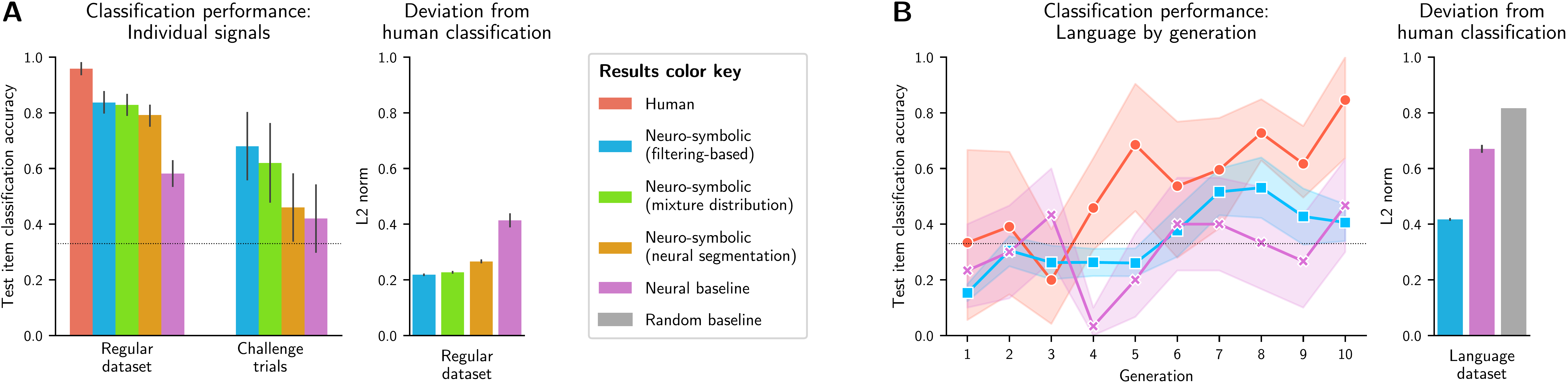}
    \caption{
    (A) The \textsc{ns} model is more human-like and more accurate than the neural baseline in classification on both the test and the challenge trials data.
    (B) In language classification, our hierarchical \textsc{ns} model is more human-like than the baseline model.
    Like in human experiments, \textsc{ns} classification accuracy increases for tasks drawn from later, more combinatorial, generations.
    }
    \vspace{-1em} %
    \label{fig:accuracies_and_human_likeness}
\end{figure*}

\paragraph{Signal classification task.}
The first task involves classifying different instances of the same signal. 
Models were trained on a subset of 800 signals from the iterated learning experiment described earlier. Test items were selected from among the remaining signals.
For each trial, a target signal and two distractors were randomly sampled. Distractors from the same language as the target were excluded. Target reproductions were obtained from the training phase of the iterated learning experiment, in which subjects had been instructed to reproduce the target signal for learning.

The goal of classification is to identify the correct reproduction of the target among the three candidate signals. Fig.~\ref{fig:tasks}A shows several example trials.
We collected human responses from 108 participants via Amazon's Mechanical Turk platform.
By matching instances of the same signal that are not point-for-point identical but share the same overall structure, we implicitly ask models to separate knowledge about part structure from low-level instance-to-instance variability.
We hypothesize that models that represent this structure explicitly will perform better.

\paragraph{Out-of-distribution challenge trials.}
To evaluate models' ability to generalize out of distribution in a more controlled way, we curated a set of `challenge trials' consisting of targets, reproductions, and carefully designed distractors (see, Fig.~\ref{fig:tasks}A~(iv), for an example).
We expect performance on those trials to be lower for all models and model variants, but anticipate particularly poor performance for non-symbolic models. 
No human judgments were collected for this dataset.

\paragraph{Language classification task.}
To evaluate the hierarchical model we used a language-level classification task.  %
We collected data from 96 human subjects on Mechanical Turk.
Models and humans were presented with a set of four signals from a target language and three candidate signals, one of which also originated from the target language (see examples in Fig.~\ref{fig:tasks}B). Trials are constructed using languages from the iterated learning experiment, which exhibit varying degrees of combinatoriality: early languages tend to lack defining features, making it hard, if not impossible, to identify the correct target signal.
Combinatoriality tends to be more pronounced in later generations (see the example languages in Fig.~\ref{fig:iterated}B), making it easier to pick out the correct signal.
For each trial, a subset of four-plus-one target signals were selected from a language at random. Distractors were sampled with the constraint that they must come from a different transmission chain than the targets.

\paragraph{Signal reproduction task.}
Finally, while classification performance constitutes an important quantitative evaluation metric, but one hallmark of human combinatorial knowledge is that it is productive: humans have the ability to generate novel instances, signals, and even entire languages.
Hence we additionally examine the models' ability to produce faithful reproductions of unfamiliar test item signals.

\paragraph{Baseline neural predictive model.}
To better understand the benefits of structured symbolic representations in our model, we compare our \gls{NS} generative model with a neural network-based model lacking explicit symbolic structure.
A predictive \gls{S2S} model \shortcite{sutskever2014sequence} that maps from a signal to its reconstruction is trained as our baseline model.
In a low-data regime like ours, such models will likely lead to low classification performance and unrealistic reproductions because they can easily overfit to the training data.
We use an \gls{LSTM} which predicts the signal $x'$ given the conditioning signal $x$.
In encoding, the \gls{LSTM} input at each timestep comprise (i) the signal value, (ii) the one-hot encoding of whether the signal is audible or silent and (iii) the end-of-signal indicator of the conditioning signal at the current timestep.
In decoding, the \gls{LSTM} input comprise these same values of the predicted signal but at the previous timestep and the \gls{LSTM} output is mapped into the parameters of distributions over the values at the current timestep.
To prevent the network from simply copying the signal, we sample the training input-output pair from a set of same instances of the signal and perform data augmentation.

\begin{figure*}[ht]
    \centering
    \includegraphics[width=\textwidth]{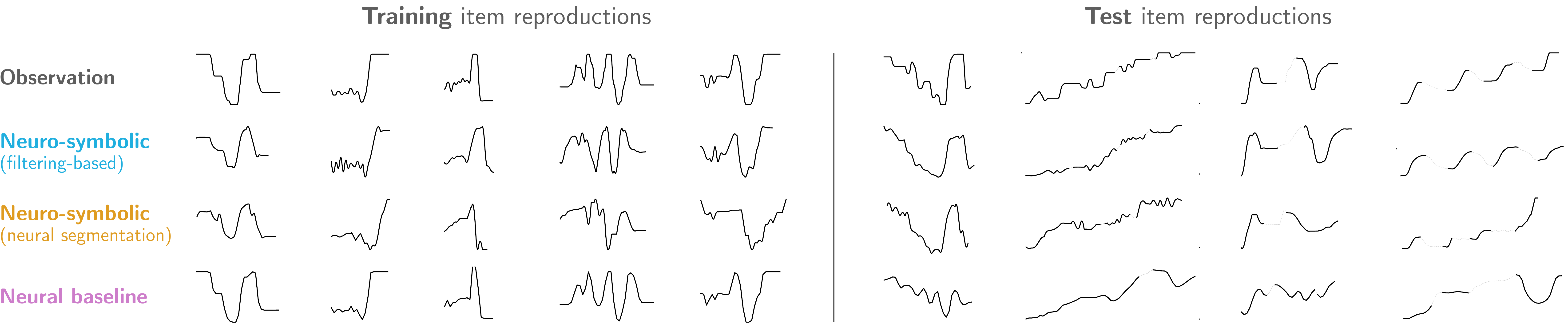}
    \caption{The \textsc{ns} generative model (both filtering- and neural segmentation- based) produces realistic reproductions of training and test items.
    The neural baseline model overfits to the training items and hence fails to accurately reproduce test items.}
    \vspace{-1em} %
    \label{fig:reconstructions}
\end{figure*}

For language-level classification, we train the neural predictive model by feeding the \gls{LSTM} encoder a subset of $M = 8$ signals and summing the embeddings before passing them to the decoder which predicts the remaining subset of signals.
We train by maximizing the predictive log probability.

In both single-signal and language classification, we select the candidate which maximizes the predictive probability.

\section{Results}
\label{sec:results}

\paragraph{Signal classification (Fig.~\ref{fig:accuracies_and_human_likeness}A).}
Our \gls{NS} model outperforms the neural baseline model on one-shot classification of test signals both in terms of ground truth accuracy and human-likeness (Fig.~\ref{fig:accuracies_and_human_likeness}A). For models and humans, we show mean performance averaged across items (trials).
For models, trial averages are obtained by rerunning training 5 times. Errorbars indicate the standard error of the mean.
For human-likeness, we show the $L_2$-norm between the classification probability $p(c \given x)$ implied by human participants and by the model predictive posteriors $p(c \given x) \propto p(x'_c \given x)$.
$p(x'_c \given x)$ is evaluated directly in the neural predictive model.
For the \gls{NS} model, we use a set of $K=10$ importance-weighted samples $(z_k, w_k)$ to approximate $p(x'_c \given x) \approx \sum_{k = 1}^K w_k p(x'_c \given z_k)$.

To understand the role of different segmentation approaches in inference, we also compare models with a strictly filtering-based vs neural segmentation.
Empirically, we find no difference in classification performance between the the filtering-based and the neural recognition models on the regular dataset ($0.83 \pm 0.35$ and $0.78 \pm 0.37$ (mean $\pm$ s.d.), respectively; $t(4)=-1.04, p=0.29$).

On the challenge trials, performance drops for all models as predicted but our model still outperforms the baseline. %

\paragraph{Language classification (Fig.~\ref{fig:accuracies_and_human_likeness}B).}
Human subject results indicate that this is a very difficult task overall, with performance starting out at chance and approaching that of the single signal task only for later generations (slope$=0.055$ $p=0.0001$). 
Like humans, the classification accuracy of our model increases when classifying languages from later generations  (slope$=0.032$ $p=0.002$) while this is not true for the baseline neural model (slope$=0.016$ $p=0.38$) (Fig.~\ref{fig:accuracies_and_human_likeness}B, left).
Our model also aligns significantly better with human categorization choices in terms of the $L_2$-norm metric introduced in single signal classification (Fig.~\ref{fig:accuracies_and_human_likeness}B, right).
Classification performance of neither model is significantly different from chance when averaged across generations.

Besides task difficulty, we note that models are trained on very few languages which come from both early and late generations of iterated learning.
Languages from early generations are less combinatorial~\shortcite{hofer2019iconicity} and hence it is might be harder for the models to learn to recognize high level patterns, providing additional noise to the already data-scarce training.
While only a first step toward modeling higher-level structure, these results suggest that a symbolic representation may allow us to better model combinatorial generalizations at the level of a language.

\paragraph{Signal reproductions (Fig.~\ref{fig:reconstructions}).}
To help convey a better qualitative sense of the models' generalization behavior, we show reproductions for signals from the training and the test set.
On both, the \gls{NS} model infers an underlying latent sequence and then based on that parse samples reproductions with variation coming from the neural renderer, leading to realistic reproductions. 
The neural predictive model does not have such explicit symbolic representation and, while `copying' the training data perfectly, produces qualitatively less realistic reconstructions on the test data.
In particular, the model appears to fail to maintain high-level discrete structure such as ``wiggly down, wiggly up'' or ``chirp and pause, repeated four times, while generally sloping upward'', which is relatively easy to discern for the \textsc{ns} model.

\section{Discussion and Outlook}
\label{sec:discussion}
The ability to acquire and propagate structured systems of knowledge such as human language is crucial for human culture. 
To study the cognitive foundations of this ability, we 
introduced a dataset of structured signals that were culturally-evolved using a slide whistle instrument, and a  neuro-symbolic generative modeling approach 
that uses representations structurally analogous to naturally occurring systems such as human phonology.
Our model's classification and signal reconstruction results, outperforming a neural baseline model, show how it begins to account for how humans understand these symbols.

One current limitation of our approach is due to trade-offs between (unsupervised) learning and model expressiveness. 
Because the \gls{LSTM}-based renderer is so expressive, it is difficult to learn an interpretable, factorized clustering of the input modality in a fully unsupervised way.

Another notable feature of the symbols we study is that they have evolved through the accumulation of multiple learning and subsequent transmission events. Improving our model to better account for the acquisition of symbols in the context of a larger hierarchical system, and extending the model to study the evolution of symbol systems over time are important next steps.

\bibliographystyle{apacite}

\setlength{\bibleftmargin}{.125in}
\setlength{\bibindent}{-\bibleftmargin}
\bibliography{main}

\begin{thebibliography}{}

\bibitem [\protect \citeauthoryear {%
de Boer%
}{%
de Boer%
}{%
{\protect \APACyear {2005}}%
}]{%
de2005evolution}
\APACinsertmetastar {%
de2005evolution}%
\begin{APACrefauthors}%
de Boer, B.%
\end{APACrefauthors}%
\unskip\
\newblock
\APACrefYearMonthDay{2005}{}{}.
\newblock
{\BBOQ}\APACrefatitle {Evolution of speech and its acquisition} {Evolution of
  speech and its acquisition}.{\BBCQ}
\newblock
\APACjournalVolNumPages{Adaptive Behavior}{13}{4}{281--292}.
\PrintBackRefs{\CurrentBib}

\bibitem [\protect \citeauthoryear {%
Feinman%
\ \BBA {} Lake%
}{%
Feinman%
\ \BBA {} Lake%
}{%
{\protect \APACyear {2020}}%
}]{%
feinman2020generating}
\APACinsertmetastar {%
feinman2020generating}%
\begin{APACrefauthors}%
Feinman, R.%
\BCBT {}\ \BBA {} Lake, B\BPBI M.%
\end{APACrefauthors}%
\unskip\
\newblock
\APACrefYearMonthDay{2020}{}{}.
\newblock
{\BBOQ}\APACrefatitle {Generating new concepts with hybrid neuro-symbolic
  models} {Generating new concepts with hybrid neuro-symbolic models}.{\BBCQ}
\newblock
\APACjournalVolNumPages{arXiv:2003.08978}{}{}{}.
\PrintBackRefs{\CurrentBib}

\bibitem [\protect \citeauthoryear {%
Ganin%
, Kulkarni%
, Babuschkin%
, Eslami%
\BCBL {}\ \BBA {} Vinyals%
}{%
Ganin%
\ \protect \BOthers {.}}{%
{\protect \APACyear {2018}}%
}]{%
ganin2018synthesizing}
\APACinsertmetastar {%
ganin2018synthesizing}%
\begin{APACrefauthors}%
Ganin, Y.%
, Kulkarni, T.%
, Babuschkin, I.%
, Eslami, S\BPBI A.%
\BCBL {}\ \BBA {} Vinyals, O.%
\end{APACrefauthors}%
\unskip\
\newblock
\APACrefYearMonthDay{2018}{}{}.
\newblock
{\BBOQ}\APACrefatitle {Synthesizing Programs for Images using Reinforced
  Adversarial Learning} {Synthesizing programs for images using reinforced
  adversarial learning}.{\BBCQ}
\newblock
\BIn{} \APACrefbtitle {{ICML}.} {{ICML}.}
\PrintBackRefs{\CurrentBib}

\bibitem [\protect \citeauthoryear {%
Hewitt%
, Le%
\BCBL {}\ \BBA {} Tenenbaum%
}{%
Hewitt%
\ \protect \BOthers {.}}{%
{\protect \APACyear {2020}}%
}]{%
hewitt2020learning}
\APACinsertmetastar {%
hewitt2020learning}%
\begin{APACrefauthors}%
Hewitt, L\BPBI B.%
, Le, T\BPBI A.%
\BCBL {}\ \BBA {} Tenenbaum, J\BPBI B.%
\end{APACrefauthors}%
\unskip\
\newblock
\APACrefYearMonthDay{2020}{}{}.
\newblock
{\BBOQ}\APACrefatitle {Learning to learn generative programs with Memoised
  Wake-Sleep} {Learning to learn generative programs with memoised
  wake-sleep}.{\BBCQ}
\newblock
\BIn{} \APACrefbtitle {{UAI}.} {{UAI}.}
\PrintBackRefs{\CurrentBib}

\bibitem [\protect \citeauthoryear {%
Hinton%
, Dayan%
, Frey%
\BCBL {}\ \BBA {} Neal%
}{%
Hinton%
\ \protect \BOthers {.}}{%
{\protect \APACyear {1995}}%
}]{%
hinton1995wake}
\APACinsertmetastar {%
hinton1995wake}%
\begin{APACrefauthors}%
Hinton, G\BPBI E.%
, Dayan, P.%
, Frey, B\BPBI J.%
\BCBL {}\ \BBA {} Neal, R\BPBI M.%
\end{APACrefauthors}%
\unskip\
\newblock
\APACrefYearMonthDay{1995}{}{}.
\newblock
{\BBOQ}\APACrefatitle {The ``wake-sleep" algorithm for unsupervised neural
  networks} {The ``wake-sleep" algorithm for unsupervised neural
  networks}.{\BBCQ}
\newblock
\APACjournalVolNumPages{Science}{268}{5214}{1158--1161}.
\PrintBackRefs{\CurrentBib}

\bibitem [\protect \citeauthoryear {%
Hofer%
\ \BBA {} Levy%
}{%
Hofer%
\ \BBA {} Levy%
}{%
{\protect \APACyear {2019}}%
}]{%
hofer2019iconicity}
\APACinsertmetastar {%
hofer2019iconicity}%
\begin{APACrefauthors}%
Hofer, M.%
\BCBT {}\ \BBA {} Levy, R.%
\end{APACrefauthors}%
\unskip\
\newblock
\APACrefYearMonthDay{2019}{}{}.
\newblock
{\BBOQ}\APACrefatitle {Iconicity and Structure in the Emergence of
  Combinatoriality} {Iconicity and structure in the emergence of
  combinatoriality}.{\BBCQ}
\newblock
\BIn{} \APACrefbtitle {Proceedings of the 41st Annual Meeting of the Cognitive
  Science Society.} {Proceedings of the 41st annual meeting of the cognitive
  science society.}
\PrintBackRefs{\CurrentBib}

\bibitem [\protect \citeauthoryear {%
Kingma%
\ \BBA {} Welling%
}{%
Kingma%
\ \BBA {} Welling%
}{%
{\protect \APACyear {2014}}%
}]{%
kingma2014auto}
\APACinsertmetastar {%
kingma2014auto}%
\begin{APACrefauthors}%
Kingma, D\BPBI P.%
\BCBT {}\ \BBA {} Welling, M.%
\end{APACrefauthors}%
\unskip\
\newblock
\APACrefYearMonthDay{2014}{}{}.
\newblock
{\BBOQ}\APACrefatitle {Auto-encoding variational {Bayes}} {Auto-encoding
  variational {Bayes}}.{\BBCQ}
\newblock
\BIn{} \APACrefbtitle {{ICLR}.} {{ICLR}.}
\PrintBackRefs{\CurrentBib}

\bibitem [\protect \citeauthoryear {%
Kirby%
, Griffiths%
\BCBL {}\ \BBA {} Smith%
}{%
Kirby%
\ \protect \BOthers {.}}{%
{\protect \APACyear {2014}}%
}]{%
kirby2014iterated}
\APACinsertmetastar {%
kirby2014iterated}%
\begin{APACrefauthors}%
Kirby, S.%
, Griffiths, T.%
\BCBL {}\ \BBA {} Smith, K.%
\end{APACrefauthors}%
\unskip\
\newblock
\APACrefYearMonthDay{2014}{}{}.
\newblock
{\BBOQ}\APACrefatitle {Iterated learning and the evolution of language}
  {Iterated learning and the evolution of language}.{\BBCQ}
\newblock
\APACjournalVolNumPages{Current opinion in neurobiology}{28}{}{108--114}.
\PrintBackRefs{\CurrentBib}

\bibitem [\protect \citeauthoryear {%
Kirby%
, Tamariz%
, Cornish%
\BCBL {}\ \BBA {} Smith%
}{%
Kirby%
\ \protect \BOthers {.}}{%
{\protect \APACyear {2015}}%
}]{%
kirby2015compression}
\APACinsertmetastar {%
kirby2015compression}%
\begin{APACrefauthors}%
Kirby, S.%
, Tamariz, M.%
, Cornish, H.%
\BCBL {}\ \BBA {} Smith, K.%
\end{APACrefauthors}%
\unskip\
\newblock
\APACrefYearMonthDay{2015}{}{}.
\newblock
{\BBOQ}\APACrefatitle {Compression and communication in the cultural evolution
  of linguistic structure} {Compression and communication in the cultural
  evolution of linguistic structure}.{\BBCQ}
\newblock
\APACjournalVolNumPages{Cognition}{141}{}{87--102}.
\PrintBackRefs{\CurrentBib}

\bibitem [\protect \citeauthoryear {%
Lake%
, Salakhutdinov%
\BCBL {}\ \BBA {} Tenenbaum%
}{%
Lake%
\ \protect \BOthers {.}}{%
{\protect \APACyear {2015}}%
}]{%
lake2015human}
\APACinsertmetastar {%
lake2015human}%
\begin{APACrefauthors}%
Lake, B.%
, Salakhutdinov, R.%
\BCBL {}\ \BBA {} Tenenbaum, J\BPBI B.%
\end{APACrefauthors}%
\unskip\
\newblock
\APACrefYearMonthDay{2015}{}{}.
\newblock
{\BBOQ}\APACrefatitle {Human-level concept learning through probabilistic
  program induction} {Human-level concept learning through probabilistic
  program induction}.{\BBCQ}
\newblock
\APACjournalVolNumPages{Science}{350}{6266}{1332--1338}.
\PrintBackRefs{\CurrentBib}

\bibitem [\protect \citeauthoryear {%
Little%
, Ery{\i}lmaz%
\BCBL {}\ \BBA {} De~Boer%
}{%
Little%
\ \protect \BOthers {.}}{%
{\protect \APACyear {2017}}%
}]{%
little2017signal}
\APACinsertmetastar {%
little2017signal}%
\begin{APACrefauthors}%
Little, H.%
, Ery{\i}lmaz, K.%
\BCBL {}\ \BBA {} De~Boer, B.%
\end{APACrefauthors}%
\unskip\
\newblock
\APACrefYearMonthDay{2017}{}{}.
\newblock
{\BBOQ}\APACrefatitle {Signal dimensionality and the emergence of combinatorial
  structure} {Signal dimensionality and the emergence of combinatorial
  structure}.{\BBCQ}
\newblock
\APACjournalVolNumPages{Cognition}{168}{}{1--15}.
\PrintBackRefs{\CurrentBib}

\bibitem [\protect \citeauthoryear {%
Nowak%
, Krakauer%
\BCBL {}\ \BBA {} Dress%
}{%
Nowak%
\ \protect \BOthers {.}}{%
{\protect \APACyear {1999}}%
}]{%
nowak1999error}
\APACinsertmetastar {%
nowak1999error}%
\begin{APACrefauthors}%
Nowak, M\BPBI A.%
, Krakauer, D\BPBI C.%
\BCBL {}\ \BBA {} Dress, A.%
\end{APACrefauthors}%
\unskip\
\newblock
\APACrefYearMonthDay{1999}{}{}.
\newblock
{\BBOQ}\APACrefatitle {An error limit for the evolution of language} {An error
  limit for the evolution of language}.{\BBCQ}
\newblock
\APACjournalVolNumPages{Proceedings of the Royal Society of London. Series B:
  Biological Sciences}{266}{1433}{2131--2136}.
\PrintBackRefs{\CurrentBib}

\bibitem [\protect \citeauthoryear {%
Roberts%
, Lewandowski%
\BCBL {}\ \BBA {} Galantucci%
}{%
Roberts%
\ \protect \BOthers {.}}{%
{\protect \APACyear {2015}}%
}]{%
roberts2015communication}
\APACinsertmetastar {%
roberts2015communication}%
\begin{APACrefauthors}%
Roberts, G.%
, Lewandowski, J.%
\BCBL {}\ \BBA {} Galantucci, B.%
\end{APACrefauthors}%
\unskip\
\newblock
\APACrefYearMonthDay{2015}{}{}.
\newblock
{\BBOQ}\APACrefatitle {How communication changes when we cannot mime the world:
  Experimental evidence for the effect of iconicity on combinatoriality} {How
  communication changes when we cannot mime the world: Experimental evidence
  for the effect of iconicity on combinatoriality}.{\BBCQ}
\newblock
\APACjournalVolNumPages{Cognition}{141}{}{52--66}.
\PrintBackRefs{\CurrentBib}

\bibitem [\protect \citeauthoryear {%
Senghas%
, Kita%
\BCBL {}\ \BBA {} {\"O}zy{\"u}rek%
}{%
Senghas%
\ \protect \BOthers {.}}{%
{\protect \APACyear {2004}}%
}]{%
senghas2004children}
\APACinsertmetastar {%
senghas2004children}%
\begin{APACrefauthors}%
Senghas, A.%
, Kita, S.%
\BCBL {}\ \BBA {} {\"O}zy{\"u}rek, A.%
\end{APACrefauthors}%
\unskip\
\newblock
\APACrefYearMonthDay{2004}{}{}.
\newblock
{\BBOQ}\APACrefatitle {Children creating core properties of language: Evidence
  from an emerging sign language in {Nicaragua}} {Children creating core
  properties of language: Evidence from an emerging sign language in
  {Nicaragua}}.{\BBCQ}
\newblock
\APACjournalVolNumPages{Science}{305}{5691}{}.
\PrintBackRefs{\CurrentBib}

\bibitem [\protect \citeauthoryear {%
Smith%
, Tamariz%
\BCBL {}\ \BBA {} Kirby%
}{%
Smith%
\ \protect \BOthers {.}}{%
{\protect \APACyear {2013}}%
}]{%
smith2013linguistic}
\APACinsertmetastar {%
smith2013linguistic}%
\begin{APACrefauthors}%
Smith, K.%
, Tamariz, M.%
\BCBL {}\ \BBA {} Kirby, S.%
\end{APACrefauthors}%
\unskip\
\newblock
\APACrefYearMonthDay{2013}{}{}.
\newblock
{\BBOQ}\APACrefatitle {Linguistic structure is an evolutionary trade-off
  between simplicity and expressivity} {Linguistic structure is an evolutionary
  trade-off between simplicity and expressivity}.{\BBCQ}
\newblock
\BIn{} \APACrefbtitle {Proceedings of the Annual Meeting of the Cognitive
  Science Society} {Proceedings of the annual meeting of the cognitive science
  society}\ (\BVOL~35).
\PrintBackRefs{\CurrentBib}

\bibitem [\protect \citeauthoryear {%
Sutskever%
, Vinyals%
\BCBL {}\ \BBA {} Le%
}{%
Sutskever%
\ \protect \BOthers {.}}{%
{\protect \APACyear {2014}}%
}]{%
sutskever2014sequence}
\APACinsertmetastar {%
sutskever2014sequence}%
\begin{APACrefauthors}%
Sutskever, I.%
, Vinyals, O.%
\BCBL {}\ \BBA {} Le, Q\BPBI V.%
\end{APACrefauthors}%
\unskip\
\newblock
\APACrefYearMonthDay{2014}{}{}.
\newblock
{\BBOQ}\APACrefatitle {Sequence to sequence learning with neural networks}
  {Sequence to sequence learning with neural networks}.{\BBCQ}
\newblock
\BIn{} \APACrefbtitle {{NeurIPS}.} {{NeurIPS}.}
\PrintBackRefs{\CurrentBib}

\bibitem [\protect \citeauthoryear {%
Verhoef%
, Kirby%
\BCBL {}\ \BBA {} De~Boer%
}{%
Verhoef%
\ \protect \BOthers {.}}{%
{\protect \APACyear {2014}}%
}]{%
verhoef2014emergence}
\APACinsertmetastar {%
verhoef2014emergence}%
\begin{APACrefauthors}%
Verhoef, T.%
, Kirby, S.%
\BCBL {}\ \BBA {} De~Boer, B.%
\end{APACrefauthors}%
\unskip\
\newblock
\APACrefYearMonthDay{2014}{}{}.
\newblock
{\BBOQ}\APACrefatitle {Emergence of combinatorial structure and economy through
  iterated learning with continuous acoustic signals} {Emergence of
  combinatorial structure and economy through iterated learning with continuous
  acoustic signals}.{\BBCQ}
\newblock
\APACjournalVolNumPages{Journal of Phonetics}{43}{}{57--68}.
\PrintBackRefs{\CurrentBib}

\bibitem [\protect \citeauthoryear {%
Witkin%
}{%
Witkin%
}{%
{\protect \APACyear {1984}}%
}]{%
witkin1984scale}
\APACinsertmetastar {%
witkin1984scale}%
\begin{APACrefauthors}%
Witkin, A.%
\end{APACrefauthors}%
\unskip\
\newblock
\APACrefYearMonthDay{1984}{}{}.
\newblock
{\BBOQ}\APACrefatitle {Scale-space filtering: A new approach to multi-scale
  description} {Scale-space filtering: A new approach to multi-scale
  description}.{\BBCQ}
\newblock
\BIn{} \APACrefbtitle {{ICASSP}} {{ICASSP}}\ (\BVOL~9, \BPGS\ 150--153).
\PrintBackRefs{\CurrentBib}

\end{thebibliography}

\end{document}